\documentclass[runningheads]{llncs}

 \usepackage[mobile]{eccv}
\usepackage[symbol]{footmisc}

\usepackage{eccvabbrv}

\usepackage{graphicx}
\usepackage{booktabs}

\usepackage[accsupp]{axessibility}  %

\usepackage[pagebackref,breaklinks,colorlinks,citecolor=eccvblue]{hyperref}

\usepackage{orcidlink}

\begin{document}

\title{4DGen: Grounded 4D Content Generation with Spatial-temporal Consistency} 

\titlerunning{4DGen: Grounded 4D Content Generation}

\author{%
  Yuyang Yin\textsuperscript{1*}, Dejia Xu\textsuperscript{2*}, Zhangyang Wang\textsuperscript{2}, Yao Zhao\textsuperscript{1}, Yunchao Wei\textsuperscript{1\textdagger}
}

\authorrunning{Y. Yin, D. Xu et al.}

\institute{\textsuperscript{1}Beijing Jiaotong University,\textsuperscript{2}
University of Texas at Austin
\\
 {\tt\small yuyangyin@bjtu.edu.cn, dejia@utexas.edu}}

\maketitle

\begin{figure}[h]
     \vspace{-1mm}
    \centering
    \includegraphics[width=0.99\textwidth]{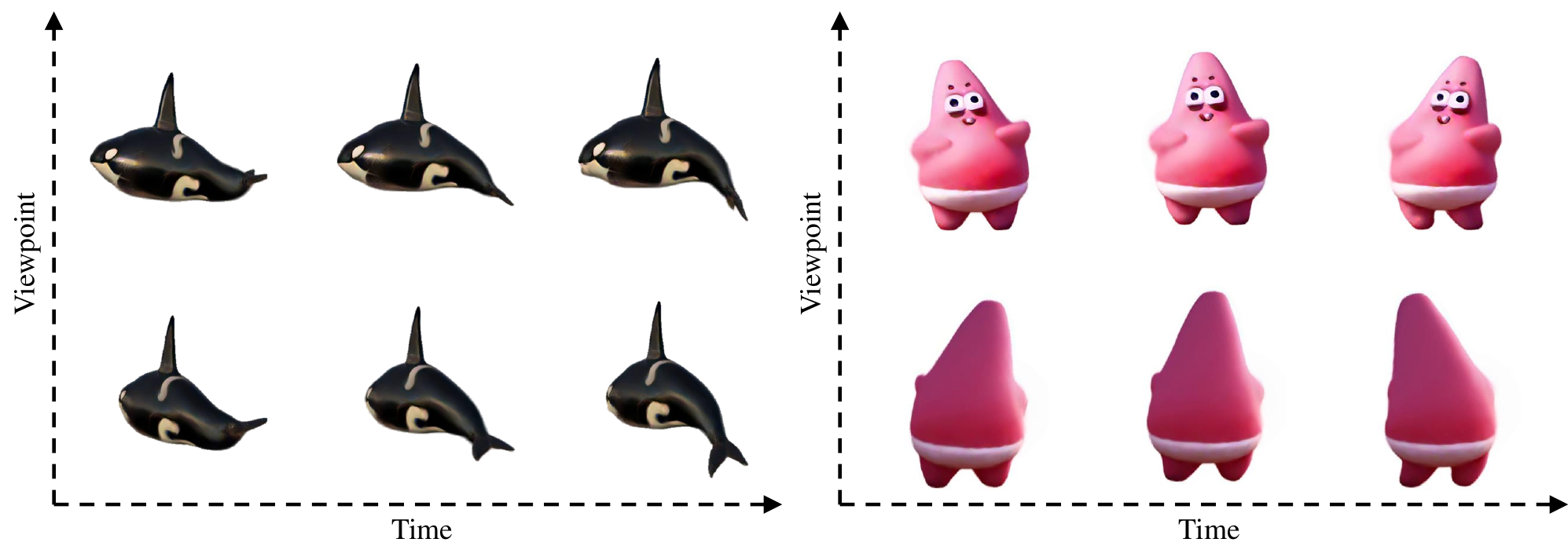}    
\caption{4DGen introduces grounded 4D content generation. We present high-quality rendered images from diverse viewpoints at distinct timesteps. Our model exhibits rapid training and high-quality results with outstanding spatial-temporal consistency.}
\label{orca}
\vspace{-5mm}
\end{figure}
\renewcommand{\thefootnote}{\fnsymbol{footnote}}
\footnotetext[1]{Equal Contribution.}
\footnotetext[2]{Corresponding Author.}
\renewcommand{\thefootnote}{\arabic{footnote}}
\begin{abstract}
 Aided by text-to-image and text-to-video diffusion models, existing 4D content creation pipelines utilize score distillation sampling to optimize the entire dynamic 3D scene. However, as these pipelines generate 4D content from text or image inputs directly, 
 they are constrained by limited motion capabilities and depend on unreliable prompt engineering for desired results.
 To address these problems, this work introduces \textbf{4DGen}, a novel framework for grounded 4D content creation. We identify monocular video sequences as a key component in constructing the 4D content.
Our pipeline facilitates controllable 4D generation, enabling users to specify the motion via monocular video or adopt image-to-video generations, thus offering superior control over content creation.
Furthermore, we construct our 4D representation using dynamic 3D Gaussians, which permits efficient, high-resolution supervision through rendering during training, thereby facilitating high-quality 4D generation.
Additionally, we employ spatial-temporal pseudo labels on anchor frames, along with seamless consistency priors implemented through 3D-aware score distillation sampling and smoothness regularizations.
Compared to existing video-to-4D baselines, our approach yields superior results in faithfully reconstructing input signals and realistically inferring renderings from novel viewpoints and timesteps.
More importantly, compared to previous image-to-4D and text-to-4D works, 4DGen supports grounded generation, offering users enhanced control and improved motion generation capabilities, a feature difficult to achieve with previous methods. Project page: \url{https://vita-group.github.io/4DGen/}
\end{abstract}

\section{Introduction}
\label{sec:intro}

The rapid advancements of the recent text-to-image diffusion models~\cite{rombach2022high, nichol2021glide, ramesh2022hierarchical, imagen} have introduced a new generative AI era. Alongside generating images from text prompts, much attention has been put into generating more complex content, such as videos and (dynamic) 3D assets. Modern artists have previously been relying on special software tools to implement their ideas into reality. As a result, automatic content creation pipelines are in great need to effectively assist human labor and reduce time-consuming manual adjustments.

Aiming at these ambitious goals, much effort has been made into creating 3D objects and videos from images or text prompts~\cite{poole2022dreamfusion,watson2022novel,liu2023one,liu2023zero,khachatryan2023text2video,ho2022imagen,singer2022make,villegas2022phenaki}. 
Researchers have shown that 3D and video datasets~\cite{deitke2023objaverse,wu2023omniobject3d,yu2023mvimgnet,wang2023internvid,Bain21} provide rich domain-specific knowledge that is beneficial in building large generative models. 
Optimization-based 3D generation also attracts great attention, relying more on 2D priors and reducing the need for 3D annotations. DreamFusion~\cite{poole2022dreamfusion} introduces score distillation sampling (SDS) by distilling knowledge from pre-trained 2D diffusion models without the need for any 3D annotation. 3D geometry and appearance are obtained by fusing multi-view supervision into a single 3D representation, implemented with Neural Radiance Field (NeRF)~\cite{xu2022neurallift,melas2023realfusion,wang2023score,seo2023let}, Mesh~\cite{wang2023prolificdreamer,lin2023magic3d}, and 3D Gaussians~\cite{chen2023text,tang2023dreamgaussian}.

Despite the exciting progress on either 3D or video generation, little attention has been put to the intersection of these two directions, \textit{i.e.} the generation of 4D (dynamic 3D) assets, like Fig.~\ref{orca}, mainly due to the lack of high-quality data.
Existing approaches mainly focus on category-specific generation~\cite{xu2018monoperfcap,habermann2020deepcap,jakab2023farm3d,zhang2023seeing,huang2024tech}.
The most related work for general object generation is MAV3D~\cite{mav3d}, where SDS is applied on text-to-image and text-to-video diffusion models to distill a 4D representation from scratch. 
Consistent4D~\cite{jiang2023consistent4d} uses interpolation-driven consistency loss and per-frame SDS to lift a monocular video sequence to 4D. 
However, their results suffer from multiple issues.
\textbf{Firstly}, MAV3D's reliance on ambiguous single image or textual inputs does not reliably yield desired 4D outcomes with coherent motion, often resulting in limited motion scope and necessitating extensive prompt engineering, thus increasing trial and error costs in practical scenarios.
\textbf{Secondly}, the Hexplane~\cite{cao2023hexplane} representation, despite its efficiency over traditional NeRF approaches, struggles with memory-intensive, high-resolution rendering over extended frames, compelling the use of lower resolution in score distillation and leading to artifacts.
Though MAV3D~\cite{mav3d} and Consistent4D~\cite{jiang2023consistent4d} attempt to mitigate these issues through super-resolution models, the outcomes frequently exhibit blurred textures and inconsistent geometry, detracting from the overall quality of the generated 4D assets.

\begin{figure*}[t]
    \centering
    \includegraphics[width=1\textwidth]{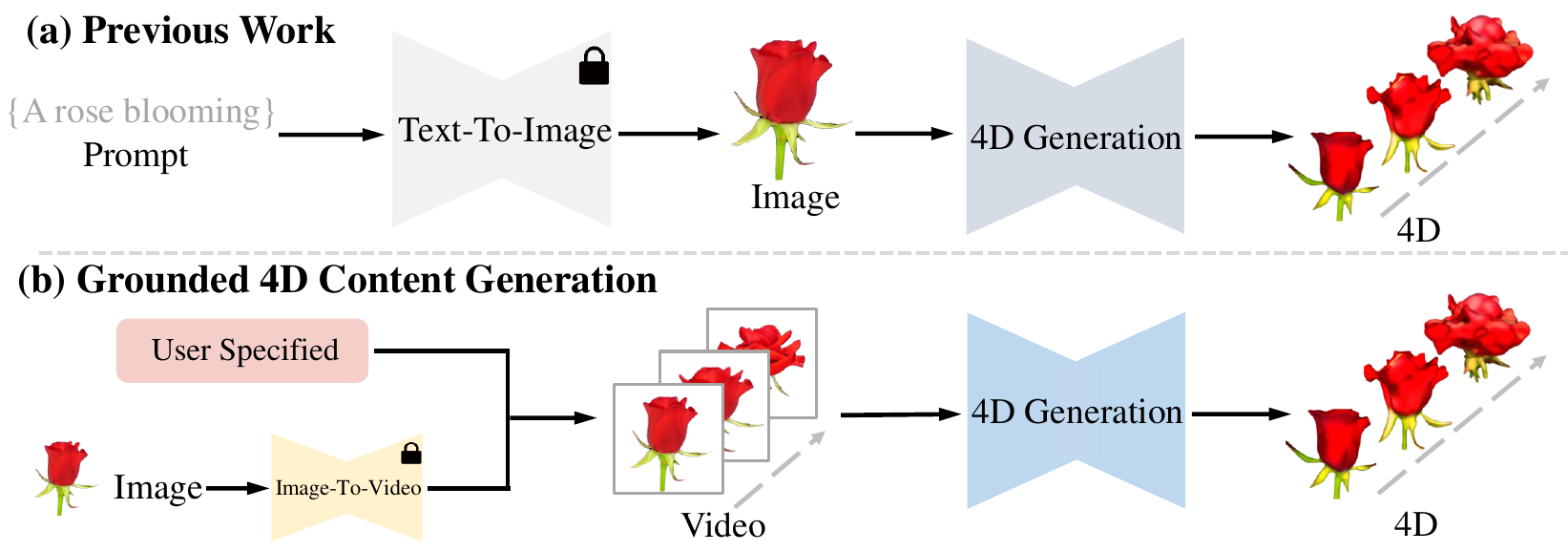}
    \caption{Previous work generates 4D content by vague single image or text input, leading to uncertain results.
    Our work introduces \textbf{Grounded 4D Content Generation}, which employs a video sequence specify the appearance and motion. Our work supports video-to-4D generation, where the input video can either be user-specified or generated from video diffusion models.
    }
    \label{fig:grounded4d}
\end{figure*}

To address the aforementioned challenges, 
we introduce \textbf{4DGen}, a novel pipeline tackling a new task of \textbf{Grounded 4D Generation}, which focuses on video-to-4D generation.
As shown in Fig.~\ref{fig:grounded4d}, our primary strategy involves using monocular videos as conditional inputs to provide users with precise control over both the motion and appearance of generated 4D content.
This strategy addresses the limited motion capabilities and controllability issues associated with previous methods.
Notably, videos can be either user-provided or synthesized through video diffusion processes. With the support of an image-to-video diffusion model~\cite{svd}, our pipeline facilitates an image-to-video-to-4D conversion.

Additionally, we implement our 4D asset with deformable 3D Gaussians, an efficient 4D representation that enables high-resolution rendering at the training stage.
Specifically, we first construct static 3D Gaussians to support deformation into motion sequences effectively. Subsequently, we employ spatial-temporal pseudo labels generated by a multi-view diffusion model and score distillation sampling loss. This approach injects motion information into the 4D representation at anchor frames. Furthermore, we use seamless spatial-temporal consistency priors to refine renderings from any viewpoint and at any timestep.

In summary, our contributions are,

\begin{itemize}
    \item We present \textbf{4DGen}, a novel pipeline for grounded 4D content generation. Our framework allows full control over the 4D asset's appearance and motion by specifying a monocular video sequence.
    \item Leveraging efficient dynamic 3D Gaussian Splatting for scene representation, our model is trained at a high resolution and long frame length, leading to visually pleasing 4D generation, as shown in Fig.~\ref{orca}.
    \item Utilizing spatial-temporal pseudo labels, we directly incorporate motion and appearance information into anchor frames. Additionally, we ensure seamless spatial-temporal consistency from any viewpoint and at any timestep through 3D-aware score distillation sampling and unsupervised smoothness regularization.	
\end{itemize}

Our experiments showcase 4DGen's superiority over per-frame generation baselines and existing 4D generation techniques in video-to-4D tasks. Our 4DGen framework delivers faithful generation of the input signals while synthesizing plausible results for novel viewpoints and timesteps. Compared with previous image-to-4D and text-to-4D works, 4DGen offers better motion generation and increased user control, surpassing the capabilities of prior methods.

\section{Related Works}
\subsection{3D Representations for Content Creation}

\looseness=-1
Since 3D data are not naturally stored in grids as pixels in 2D, multiple 3D representations have been studied for content creation.
Polygonal meshes represent shape surfaces using vertices, edges, and faces.
Previous works~\cite{gao2022get3d,shen2021deep,pavllo2021learning,pavllo2020convolutional} have achieved high quality textured 3D meshes generation.
On the other hand, point cloud utilizes an unstructured set of points in 3D space to represent surfaces. Many works have explored point cloud generation by autoencoding~\cite{achlioptas2018learning,gadelha2018multiresolution, yang2018foldingnet}, adversarial generation~\cite{shu20193d,valsesia2018learning}, and diffusion model~\cite{zeng2022lion,zhou20213d}.
However, both meshes and point clouds are known to require significant amounts of memory and slow training. NeRF ~\cite{mildenhall2021nerf} addresses this challenge by implicitly employing Multilayer Perceptrons (MLPs) to represent objects and scenes. 
While easy to optimize, rendering NeRF into high-resolution images usually requires millions of queries of the MLP network, making it hard to supervise through patch-based rendering losses. Many follow-up works aim to overcome this issue by introducing hybrid representations that leverage explicit structure to reduce the burden of NeRF's MLP.
Instant NGP~\cite{muller2022instant} significantly accelerates neural rendering using a compact neural network combined with a multi-resolution hash table. EG3D~\cite{chan2022efficient} introduces an expressive hybrid explicit-implicit network architecture for the unsupervised generation of high-quality multi-view-consistent images and 3D shapes.
Recently, 3D Gaussian Splatting (3D-GS)~\cite{kerbl20233d} introduces an alternative 3D representation to NeRF. 3D-GS demonstrates impressive visual quality and real-time rendering, supported by a rapid visibility-aware rendering algorithm.
Our work follows this inspiring direction in representing the 3D scene with 3D Gaussians, and we adopt an additional implicit network to make the 3D Gaussians deformable for dynamic scenes.

\subsection{3D Content Generation with Diffusion Prior}
\looseness=-1
Diffusion models are widely used across various fields~\cite{rombach2022high,podell2023sdxl,zhang2023controlvideo,yin2023cle,hu2025diffusion,huang2024classdiffusion}, and the generation of 3D content from multi-modal inputs has garnered significant research interest over the years.
The task of text-to-3D generation focuses on generating a 3D model from a textual prompt.
Due to the lack of large-scale 3D data, many works try to improve 3D generation quality by improving the rendering quality from random 2D viewpoints.
Early works  DreamFields~\cite{jain2022zero} and CLIPMesh~\cite{mohammad2022clip} utilized CLIP~\cite{radford2021learning}, but the results tend to lack realism. With the help of 2D diffusion priors,
the seminal work DreamFusion~\cite{poole2022dreamfusion} introduces Score Distillation Sampling (SDS) and showcases encouraging results.
Many follow-up works have then tried to improve SDS sampling results through better optimization~\cite{kim2023collaborative,huang2023dreamtime,mardani2023variational,hong2023debiasing,armandpour2023re,wang2023prolificdreamer,zhong2024dreamlcm}, crafting different 3D representations~\cite{lin2023magic3d,tang2023dreamgaussian,yi2023gaussiandreamer,chen2023text} or constructing better diffusion priors that are suitable for 3D generation~\cite{liu2023zero,qian2023magic123,seo2023let,li2023sweetdreamer}.
SDS can also be adapted to image-to-3D generation~\cite{xu2022neurallift,melas2023realfusion,deng2022nerdi,liu2023zero}. More recently, with the help of large-scale 3D asset datasets~\cite{deitke2023objaverse,wu2023omniobject3d,yu2023mvimgnet}, better generative models~\cite{liu2023zero,liu2023syncdreamer,jiang2023efficient,shi2023zero123++,liu2023one,hong2023lrm,xu2023dmv3d} are built and produce outstanding 3D asset generation ability when given a single image input.
Our work draws inspiration from both directions and proposes to use pseudo labels explicitly sampled from a large generative model to supervise anchor viewpoints and employ score distillation sampling to enforce consistency for arbitrary viewpoints.

\vspace{-3mm}
\subsection{Dynamic Content Generation}
Unlike static content generation, producing high-quality dynamic datasets remains challenging due to the lack of data~\cite{wang2023internvid,Bain21}. 
Many prior works seek to leverage existing priors learned on large-scale text-image pairs~\cite{schuhmann2022laion,kakaobrain2022coyo-700m,chen2015microsoft} and adapt to dynamic content generations.
Tune-a-Video~\cite{wu2023tune} tackles one-shot video generation through subtle architecture modifications and sub-network tuning. Text2Video-Zero~\cite{khachatryan2023text2video} introduces a training-free method for animating a pre-trained text-to-image model.  AnimateDiff~\cite{guo2023animatediff} converts existing text-to-image diffusion models to video generators by designing a motion modeling module that is zero-shot transferrable to unseen stable diffusion variants. 

Regarding 4D content generation, early works mainly focus on category-specific generation, such as digital humans~\cite{xu2018monoperfcap,habermann2020deepcap,huang2024tech,loper2023smpl,liao2023tada,wang2023learning,zhang2023learning,rempe2023trace}, animals~\cite{jakab2023farm3d} and flowers~\cite{zhang2023seeing}.
Regarding general object generation, MAV3D~\cite{mav3d} leverages score distillation sampling from both image and video diffusion models to optimize a 4D scene.
Consistent4D~\cite{jiang2023consistent4d} is a concurrent method that also studies lifting a monocular video sequence to 4D. Their work leverages frame interpolation-driven consistency loss and per-frame SDS from a 3D-aware diffusion model. 
Animate124~\cite{zhao2023animate124} and 4D-fy~\cite{bahmani20234d} are concurrent methods achieve image-to-4D or text-to-4D by video SDS loss.
Our work similarly studies utilizing 3D-aware diffusion priors for 4D generation, but differs in that we not only employ SDS supervision but also introduce DDIM-sampled pseudo labels for supervision at a higher resolution. Moreover, while their work adopts costly NeRF-based volumetric rendering, we utilize an efficient rasterizer brought by 3D Gaussians, supporting faster training implemented through rendering losses.

\section{4DGen: Grounded 4D Generation}
\label{sec:method}

Generating 4D contents without 4D annotation is highly ill-posed since multiple 4D results can all be plausible when projecting to a monocular video sequence. 
To this end, we adopt a pre-trained 3D-aware multi-view diffusion model to provide spatial-temporal pseudo labels.
Additionally, since the spatial-temporal pseudo labels are only provided on limited 2D renderings, we include score distillation sampling from a 3D-aware diffusion model to supervise random viewpoints and unsupervised smoothness regularization to enforce temporal smoothness at random timestep in 3D explicitly.

\begin{figure*}[t]
    \centering
    \includegraphics[width=1\textwidth]{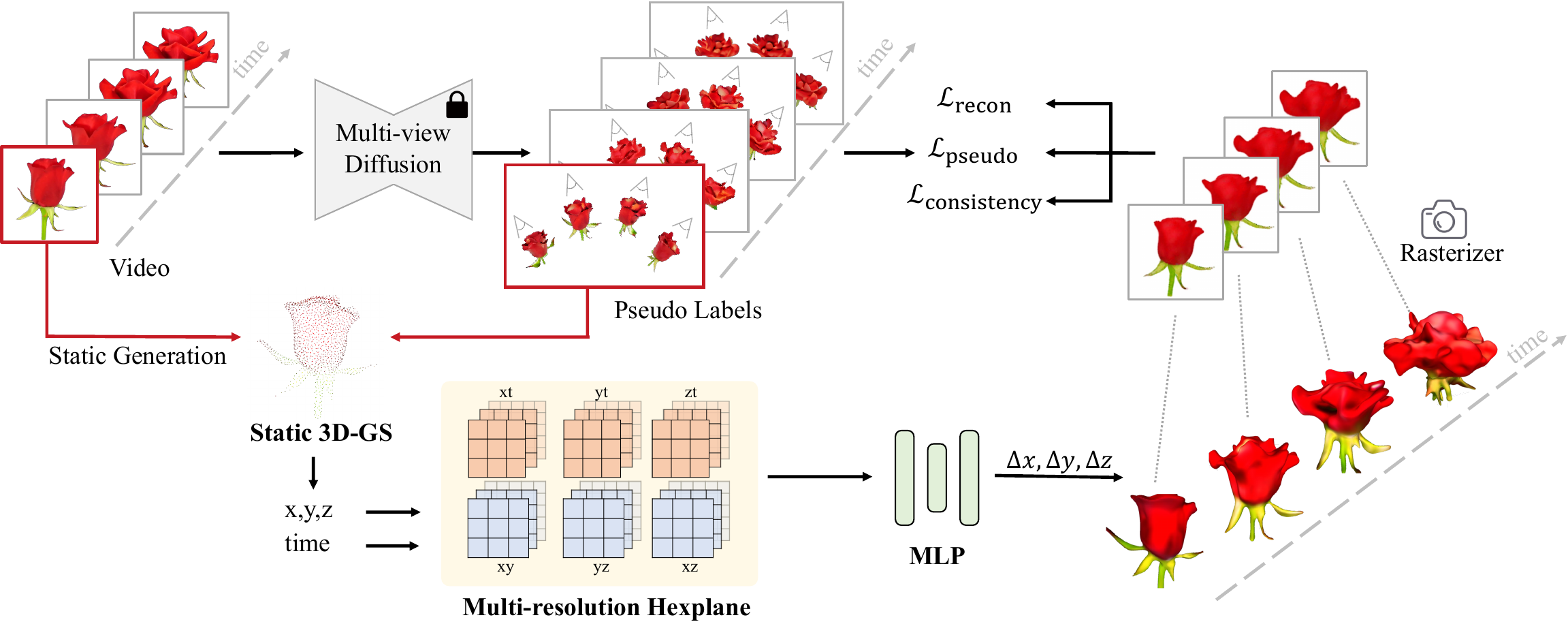}
    \caption{Overall framework of 4DGen. We conduct 4D generation grounded by a static 3D asset and a monocular video sequence.
    Our 4D scene is implemented by deforming a static set of 3D Gaussians. Besides reconstructing the input signals, we supervise our framework with spatial-temporal pseudo labels on anchor frames by a multi-view diffusion model.
    Moreover, we adopt seamless consistency priors implemented with score distillation sampling and unsupervised smoothness regularizations. 
    }
    \label{fig:framework}
\end{figure*}

In this section, we first introduce the background of score distillation sampling and 3D Gaussian splatting. Then, we construct our scene by generating a static scene via 3D Gaussians. Subsequently, we introduce our spatial-temporal pseudo labels on anchoring viewpoints. Finally, we propose seamless spatial and temporal consistency priors for arbitrary viewpoints and timesteps.
An overview of our framework is provided in Fig.~\ref{fig:framework}. 

\subsection{Preliminary}

\subsubsection{Score Distillation Sampling}

Score Distillation Sampling (SDS) is widely adopted for distilling 2D image priors from a pre-trained diffusion model $\epsilon_{\phi}$ into 3D representations through optimization on the renderings. When provided with a 3D model parameterized by $\theta$ and an image generator $g$, the multi-view image $x$ can be rendered by $g(\theta)$.
SDS optimizes the 3D parameter $\theta$ as follows:
\begin{equation}
\nabla_{\theta}\mathcal{L}_\text{SDS}(\phi,x)=w(t)\left[(\hat{\epsilon_{\phi}}(x_t;y,t)-\epsilon) \right]\frac{\partial x}{\partial \theta},
\end{equation}
where $w(t)$ is a weighting function, $x_t$ is obtained by perturbing $x$ with randomly sampled Gaussian noise, and $y$ is the conditional signal. In practice, $\hat{\epsilon_{\phi}}(z_t;y,t)$ is implemented by classifier-free guidance, which combines the scores of a conditional model ${\epsilon_{\phi}}(z_t;y,t)$ and an unconditional model ${\epsilon_{\phi}}(z_t;\empty,t)$ to achieve better generation conditioned on $y$.

\subsubsection{3D Gaussian Splatting}
3D Gaussian Splatting(3D-GS)~\cite{kerbl20233d} is an explicit 3D scene representation utilizing a set of differentiable 3D Gaussians.  Each Gaussian is defined with center position $\mu \in \mathbb{R}^3$, covariance matrix $\Sigma\in \mathbb{R}^{3\times3}$, color $c \in \mathbb{R}^3$ and opacity $\alpha \in \mathbb{R}^1$. It can be formulated as:
\begin{equation}
G(x)=e^{-\frac{1}{2}{(x)}^{T}\Sigma^{-1}(x)},
\label{eq:3dGS}
\end{equation}
where x means the distance between the center position $\mu$ and the query point. Color can be calculated as:
\begin{equation}
C(r)=\sum_{i\in N}c_i\sigma_i\prod_{j=1}^{i-1}(1-\sigma_j), 
\label{eq:3dGS}
\end{equation}
where $\sigma_i=\alpha_iG(x_i)$ and $N$ means the number of sample Gaussian points.

\subsection{Static Gaussian Generation}
The vanilla 3D Gaussians are only designed for reconstructing static 3D scenes, and many follow-up works~\cite{luiten2023dynamic,wu20234d} extend the framework to dynamic scenes by constructing a static scene first and then learning the point deformations to support scene motion. We draw inspiration from 4D Gaussian Splatting~\cite{wu20234d} to use a set of static Gaussians that are shared across different timesteps and utilize an additional HexPlane~\cite{cao2023hexplane} representation to express the deformations for point attributes. 

Our framework initially generates the static Gaussian initialization to enable scene deformation.
Owing to the ill-posed and data-limited nature of our generative task, additional design considerations are necessary to construct a plausible static scene.
To achieve this, we utilize a multi-view diffusion model, pre-trained on 3D datasets, to infer the scene structure. 
We perform DDIM sampling using SyncDreamer~\cite{liu2023syncdreamer} to obtain multi-view predictions of the scene, and then optimize a set of static 3D Gaussians using the sampled images. Our framework adopts random initialization and optimizes via $\mathcal{L}_1$ loss, unlike using SDS loss in DreamGaussian~\cite{tang2023dreamgaussian}.
Specifically, for the initial frame, off-the-shelf multi-view reconstruction algorithms~\cite{wang2021neus,wang2023neus2} are employed to estimate coarse geometry and extract point clouds.

It should be noted that our static 3D Gaussians are not kept fixed but jointly optimized with the deformation field once the static 3D Gaussians are properly constructed. During the joint optimization phases, the static 3D Gaussians are not compelled to reconstruct the initial frame of the 4D scene. Rather, the deformation field can modify the static 3D Gaussians at each timestep. This approach additionally addresses the issue of poor geometric quality arising from inconsistencies in multi-view predictions.

\subsection{Spatial-temporal Anchor Frame Pseudo-labels }
MAV3D~\cite{mav3d} demonstrates outstanding performance by jointly distilling video and image diffusion priors. However, the lack of powerful open-source video diffusion models hinders the wide application of this direction. Possibly due to the lack of large-scale, high-quality dynamic video dataset, open-source video diffusion models usually leverage image-video joint training~\cite{khachatryan2023text2video} or transfer learning~\cite{guo2023animatediff,wu2023tune} to benefit from the knowledge learned during the foundational training of Stable diffusion~\cite{rombach2022high}. As a result, the video generation quality suffers from limited realism and, therefore, is not ideal enough for distillation-based 4D content generation~\cite{mav3d} directly. To address this challenge, we propose a novel lightweight solution by introducing anchor frames and explicitly specifying the 3D motion information at anchor frames for the 4D scene without injecting video diffusion priors. 

\subsubsection{Spatial-temporal 2D Pseudo Labels}
Owing to the scarcity of data, we endeavor to transfer knowledge from external priors. Given the monocular video sequence, we utilize SyncDreamer~\cite{liu2023syncdreamer} to synthesize multi-view images at each timestamp.
SyncDreamer is a diffusion model pre-trained on multi-view image datasets and aims to generate 3D consistent multi-view results. 
Consequently, we aim to distill this geometry information into the 4D content we generate.

However, while effective, SyncDreamer processes each frame individually, and for each frame, multiple 3D objects can be plausible due to the ill-posed nature of image-to-3D generation. 
Consequently, there is no consistency assurance in generation across frames, with this issue being particularly pronounced in synthesized back views.
Furthermore, despite SyncDreamer's 3D-aware architecture aiming to ensure geometric consistency in novel view predictions, 3D consistency across viewpoints from DDIM-sampled images remains unguaranteed.
Therefore, it is impractical to train a visually pleasing 4D representation using these supervisions directly. Instead, they serve as pseudo labels to facilitate the initial warming up of the 4D representation at anchor frames.

\subsection{Seamless Spatial-temporal Consistency Priors}

While efficient, dealing with anchor frames only will lead to flickering results if the model is inference with a different frame rate than training time.
Inspired by MAV3D~\cite{mav3d}, we utilize a random frame sampling rate at training time to provide temporal augmentation of the model. By randomly selecting a frame rate and a random starting time step, our model can render a random short sequence of continuous time frames for refinement. Since we do not have access to ground truth signals for intermediate frames and arbitrary viewpoints, we adopt external priors and unsupervised regularization as alternates.

\subsubsection{Arbitrary Viewpoint Spatial Consistency through SDS}
Motivated by DreamFusion~\cite{poole2022dreamfusion}, we adopt score distillation sampling (SDS) to overcome the multi-view inconsistency issue produced by the pseudo labels. Unlike regressing the inconsistent DDIM-sampled images, SDS optimizes the scene representation such that our rendered images maintain a high likelihood as evaluated by the pre-trained diffusion prior. With the help of recently released large-scale 3D-aware diffusion prior~\cite{deitke2023objaverse}, we implement our image-conditioned SDS through conditioning on the front view of each timestep. SDS not only overcomes the aforementioned inconsistent issue introduced by the SyncDreamer predictions but also introduces dense supervision coverage on arbitrary viewpoints, supporting better visual appearance.

\subsubsection{Arbitrary Timestep Temporal Consistency via Smoothness Regularization}
Thanks to the explicit nature of 3D Gaussians, we obtain the point locations for each timestep during rendering.
We thus enforce unsupervised smoothness priors in intermediate frame renderings. We first implement spatial total variance that is widely used in prior works~\cite{kplanes,cao2023hexplane},
\begin{equation}
\begin{aligned}
\mathcal{L}_\text{TV}(\mathbf{P}) &=\frac{1}{|C| n^2} \sum_{c, i, j}\left(\|\mathbf{P}_c^{i, j}-\mathbf{P}_c^{i-1, j}\right\|_2^2 
 +\left\|\mathbf{P}_c^{i, j}-\mathbf{P}_c^{i, j-1}\right\|_2^2),
\end{aligned}
\end{equation}
where $i, j$ are indices on the current plane $\mathbf{P}$. We further utilize temporal smoothness prior that penalizes the acceleration of individual 3D points over time,
\begin{equation}
    \mathcal{L}_\text{smooth}(\mathbf{P})=\frac{1}{|C| n^2} \sum_{c, i, t}\left\|\mathbf{P}_c^{i, t-1}-2 \mathbf{P}_c^{i, t}+\mathbf{P}_c^{i, t+1}\right\|_2^2.
\end{equation}
To further encourage temporal smoothness, we initialize the space-time planes used in our scene representation as a constant one so that the space-time features are not necessarily changed if the corresponding spatial content remains static across different timesteps.
The consistency priors can be formulated as follows, 
\begin{equation}
    \mathcal{L}_\text{consistency} = \mathcal{L}_\text{smooth} + \omega_2\mathcal{L}_\text{TV} + \omega_3\mathcal{L}_\text{SDS},
\end{equation}
leading to the overall loss function,
\begin{equation}
    \mathcal{L} = \mathcal{L}_\text{recon} + \omega_4\mathcal{L}_\text{pseudo} + \omega_5\mathcal{L}_\text{consistency},
\end{equation}
where $\omega_i$ are weighting factors and $\mathcal{L}_\text{recon}$ refers to LPIPS~\cite{zhang2018lpips} loss enforced to reconstruct the original input.

\section{Experiments}
\label{sec:exp}

\subsection{Implementation Details}

\subsubsection{Architecture}
\looseness=-1
We implement our 4D representation using 4D Gaussian Spaltting~\cite{wu20234d}.
We adopt a multi-resolution HexPlane voxel module to encode the deformation of 3D Gaussians. The six planes represent the combinations of spatial-temporal dimensions: $(x,y),(x,z),(y,z),(x,t),(y,t),(z,t)$, with the last three planes preserving the temporal information. Each queried feature is interpolated based on the nearby voxel grid features at each resolution. The initial voxel resolution is set to $[64,64]$. Four levels of upsampled resolutions are employed, each doubling the initial resolution. An additional MLP head is utilized to decode interpolated features into the final predictions of position deformation $\Delta\mathbf{X}$.

\subsubsection{Training}
Our training process comprises static, coarse, and fine stages. 
We start by constructing the static 3D Gaussians to support further deformation. 
If the user specifies a static 3D asset, we can initialize the static Gaussians with the point cloud converted from the 3D asset. Otherwise, we randomly initialize points in a unit sphere and spend 1,000 iterations to optimize the 3D Gaussians. We exclusively densify and prune 3D Gaussians in static stage.
In the coarse and fine stage, we jointly optimize the static 3D Gaussians and the deformation field.
Our coarse stage involves 1,000 iterations of optimization using $\mathcal{L}_\text{pseudo}$. As for the fine stage, we aim to enhance the visual quality while maintaining consistency in both spatial and temporal domains. To address the limitation of optimizing against inconsistent spatial-temporal pseudo labels, which tends to produce blurred results rather than accurately capturing the position and color of 3D Gaussian points, we thus introduce our seamless consistency priors to enhance overall consistency. We randomly use 10\% of iterations to sample arbitrary frame rate and starting timestep to enforce temporal consistency for arbitrary timestep sequences. For the rest 90\% iterations we randomly sample a timestep among the anchor frames and supervise the rendering of a random viewpoint through SDS implemented with Zero123~\cite{liu2023zero}.
The fine stage contains 3,000 iterations.

\subsection{Baseline Methods}
MAV3D~\cite{mav3d} is the first work preceding 4D generation for 4D content creation. A direct comparison is not feasible since their code has not been released. To evaluate our method, we instead establish two baseline approaches following the practice in MAV3D. DreamGaussian~\cite{tang2023dreamgaussian} is a state-of-the-art image-to-3D method utilizing 3D Gaussian splatting. DreamGaussian is provided with individual video frames to generate a total of $T$ 3D models, where $T$ is the total number of frames. Subsequently, we can render images from every view at each time. The second baseline involves One-2-3-45~\cite{liu2023one}, a high-quality multi-view diffusion model. Given an image at each time, One-2-3-45 can generate multi-view images and fuse them into a mesh. 
For these two baselines, 
we concatenate the $T$ results to construct a 4D model.

We also provide comparisons against a concurrent video-to-4D method Consistent4D~\cite{jiang2023consistent4d}, image-to-4D method Animate124~\cite{zhao2023animate124} and text-to-4D method 4D-fy~\cite{bahmani20234d}.
The datasets consist of three parts, with each test case containing only a reference video sequence from the front view. The first part consists of a video dataset released by Consistent4D~\cite{jiang2023consistent4d}. The second part of the dataset is collected either in the wild or from Sketchfab~\cite{sketchfab} by ourselves. The final part is generated using a pre-trained image-to-video diffusion model ~\cite{svd}.

\begin{figure*}
    \centering
    \includegraphics[width=\textwidth]{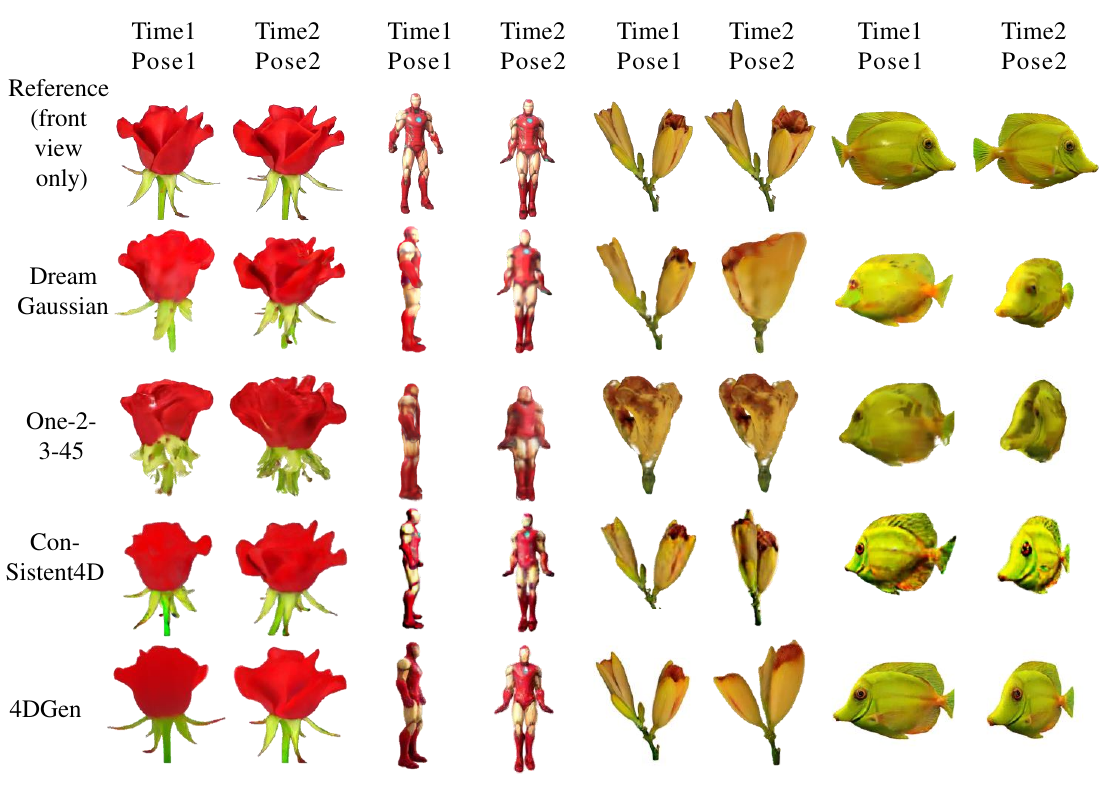}
    \caption{Comparison with baseline methods. The reference only contains front-view images, while our 4DGen model is capable of rendering at any timestep from any viewpoint. Our method maintains greater consistency and higher quality in both spatial and temporal consistency. In contrast, other methods result in significant artifacts or blurring, attributed to inaccurate feature estimation from different views. Moreover, these baselines generate individual frames separately and cannot be rendered at intermediate timesteps. We provide \textbf{more video comparisons} in the project page.
    }
    \label{fig:comparison}
\end{figure*}

\begin{figure}
\centering
\includegraphics[width=\columnwidth,keepaspectratio]{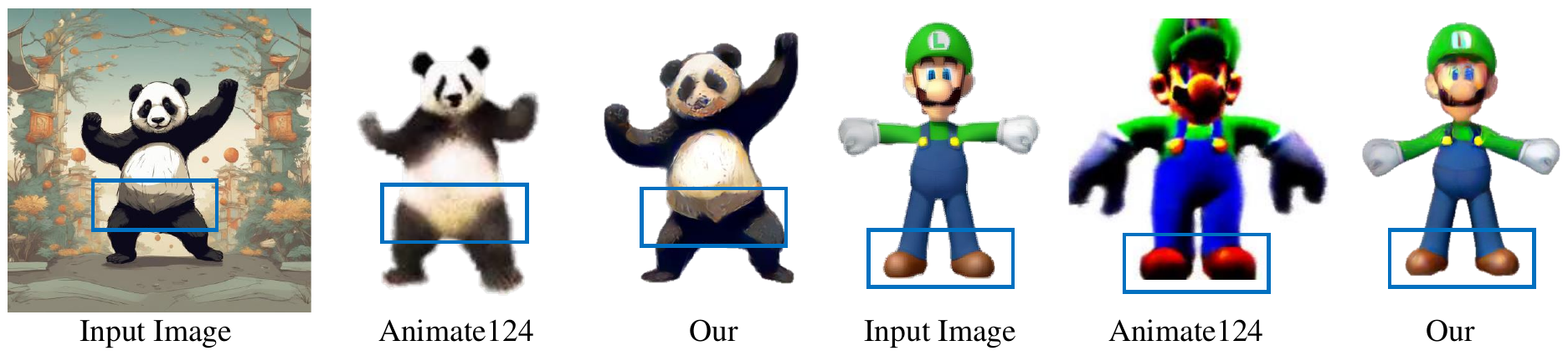}\\
\caption{
Image-to-4D Comparison.  
Our input video sequences are generated through the process image-to-video via Stable Video Diffusion~\cite{svd}. As shown in bounding box areas, our results exhibit identity consistency with the input reference image when compared to Animate124. We provide \textbf{video comparisons} in the project page.}
\label{fig:image4d}
\end{figure}

\subsection{Evaluation Metric}

Currently, there is no well-established evaluation metric for assessing the quality of 4D generations. Our assessment focuses on the quality of 4D content generation from the perspectives of space and time. In other words, images rendered from any viewpoint for any timestep should all enjoy high quality, while image sequences should demonstrate realistic consistency across frames and viewpoints. 

We use the widely adopted CLIP distance as in previous 3D content creation works~\cite{xu2022neurallift,tang2023dreamgaussian} to measure the view synthesis quality. For temporal smoothness, we employ CLIP-T as in prior works~\cite{esser2023structure,geyer2023tokenflow}, which is the average CLIP distance between adjacent frames.
Since novel view synthesis gets more challenging as novel-view camera gets more distant, we choose to also measure CLIP-T distance at different viewpoints, not only for the frontal view but also for the back and side views, denoted as CLIP-T-f, CLIP-T-b, and CLIP-T-s, respectively.  To measure the quality of videos, we use X-CLIP~\cite{ni2022expanding}, a constrastive pre-trained video-text  model from front, back and 360$^\circ$ views, denoted as XCLIP-f, XCLIP-b and XCLIP-r. The textual description is human-generated.

\begin{figure}
\centering
\includegraphics[width=\columnwidth,keepaspectratio]{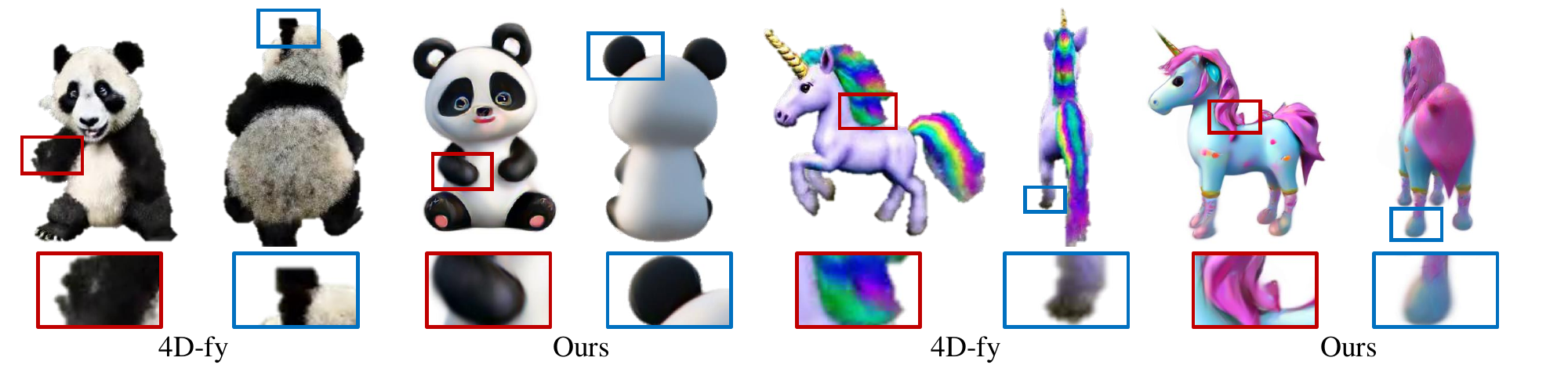}\\
\caption{
Text-to-4D Comparison.  
Our input video sequences are generated through the process of text-to-image via Stable Diffusion~\cite{rombach2022high} and image-to-video via Stable Video Diffusion~\cite{svd}. 
Our results are more reasonable in detail and there are no noticeable artifacts.
In comparison, 4D-fy results in limited motion.
We provide \textbf{video comparisons} in the project page.}
\vspace{-3mm}
\label{fig:text4d}
\end{figure}

\begin{table}[!t]
    \centering
    \vspace{-3mm}
    \caption{Average results on 13 datasets.}
    \vspace{-3mm}
    \resizebox{0.8\columnwidth}{!}{
    \begin{tabular}{lcccc}
    \toprule 
        Method & DreamGaussian & One-2-3-45 & Consistent4D& Ours \\ 
        \toprule 
        CLIP$\downarrow$ & 0.2538 & 0.3249 & 0.2508&\textbf{0.2395} \\ 
        CLIP-T-f$\downarrow$ & 0.0275 & 0.0796 & 0.0203&\textbf{0.0135} \\ 
        CLIP-T-s$\downarrow$ & 0.0647 & 0.0937 & 0.0178 &\textbf{0.0210} \\ 
        CLIP-T-b$\downarrow$ & 0.0491 & 0.0693 & -&\textbf{0.0129} \\ 
        XCLIP-f$\uparrow$ & 17.9058  & 13.8918 & 17.1490&\textbf{18.0234} \\ 
        XCLIP-b$\uparrow$ & 16.1569  & 15.2904 & 13.8882&\textbf{17.5254} \\ 
        XCLIP-r$\uparrow$ & 13.7809  & 15.9654 & -&\textbf{17.5805} \\ 
        \bottomrule
        \vspace{-5mm}
        \label{tab:comparsion}
    \end{tabular}
    }
\end{table}

\subsection{Video-to-4D Qualitative and Quantitative Comparisons}
We visualize our comparisons against baselines in Fig.~\ref{fig:comparison}, DreamGaussian~\cite{tang2023dreamgaussian} struggles to generate fine-grained details, possibly due to the limited number of Gaussian points.  One-2-3-45~\cite{liu2023one2345} adopts multi-view reconstruction from Zero123's noisy predictions. Possibly due to the limited generalization ability of Zero123, their results usually contain inaccurate geometry predictions and white texture artifacts. 
As each frame is generated individually, adapting these frameworks' results to 4D requires concatenating images from corresponding views at each time to produce a video. This results in noticeable temporal discontinuity, characterized by frequent flickering and texture inconsistencies over time. Consistent4D exhibits color distortion and artifacts, likely stemming from the limited capabilities of the NeRF model in low resolution and the post-processing stage of the super-resolution network.

Our 4DGen significantly outperforms baselines in spatial and temporal consistency, evident in enhanced details and smooth transitions between frames. Furthermore, previous methods only support the generation of a fixed number of frames, which is equal to the number of input images.
Our method, on the other hand, benefits from the seamless consistency priors, and supports sampling videos of arbitrary length from arbitrary viewpoints.
Since we utilize the output of Consistent4D~\cite{jiang2023consistent4d} as reported on its webpage, testing two metrics under identical settings is not feasible.
As shown in Table.~\ref{tab:comparsion}, we compare average results on 13 cases. Our 4DGen highly surpasses the compared methods across all metrics.

\subsection{Image/Text-to-4D Comparisons}
Our study supports input videos generated from video diffusion models, facilitating the accomplishment of image-to-4D and text-to-4D tasks. As shown in Fig.~\ref{fig:image4d}, we compare our image-to-4D results with Animate124~\cite{zhao2023animate124}. 
Animate124 utilizes a pretrained video diffusion model to driven object motion by video SDS loss. Despite employing textual inversion~\cite{gal2022image} to maintain identity with the reference image, it unavoidably leads to identity inconsistency. In contrast, our grounded 4D content generation faithfully generates objects in accordance with input signals.

Due to the flickering and inconsistent features in the current text-to-video model, we address text-to-4D challenges by employing a text-to-image model~\cite{rombach2022high,podell2023sdxl} to generate the desired front image. Subsequently, we leverage an image-to-video diffusion model~\cite{svd} to generate the desired video, feeding it into our pipeline to accomplish the text-to-4D task.
As illustrated in Fig~\ref{fig:text4d}, our model achieves better details and geometry shapes. The motion in 4D-fy~\cite{bahmani20234d} relies on video SDS loss, inevitably introducing temporal consistency issues with flickering in video results and limiting motion variability. Our results exhibit good consistency and rich motion. We provide \textbf{video comparisons} in the project page.

\subsection{Training Time and Memory}
Our model is trained for about an hour on a scene with one RTX3090 with 15G memory. DreamGaussian~\cite{tang2023dreamgaussian}  is a per-frame generation baseline, which requires 32 minutes(2 min per frame). Consistent4D~\cite{jiang2023consistent4d} runs for 2.75 hours on a V100 with 40G memory. MAV3D~\cite{mav3d} takes 6.5 hours on 8 A100s. 4D-fy~\cite{bahmani20234d} takes 23 hours on a A100. We test Animate124~\cite{zhao2023animate124} on a A6000, requiring approximately 4.5h. In comparison with the previous method, our model achieves fast training and requires less memory.

\subsection{Ablation Study}
\begin{figure}
\centering
\includegraphics[width=\columnwidth,keepaspectratio]{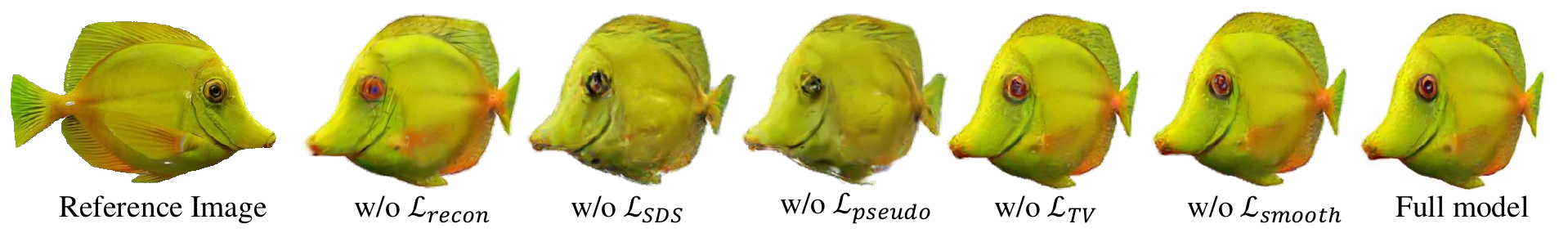}\\
\caption{Ablation studies of our proposed components. Models lacking either $\mathcal{L}_\text{recon}$, $\mathcal{L}_\text{SDS}$ or $\mathcal{L}_\text{pseudo}$ fail to generate high-quality details. Models without $\mathcal{L}_\text{TV}$ or $\mathcal{L}_\text{smooth}$ fail to generate outputs with smoothness.  We provide \textbf{video comparisons} in the project page.}
\vspace{-5mm}
\label{fig:abl}
\end{figure}

We conduct ablation studies to validate the effectiveness of our novel components on the fish dataset. As shown in Tab.~\ref{tab:abl} and Fig.~\ref{fig:abl}, when training without  $\mathcal{L}_\text{recon}$, $\mathcal{L}_\text{SDS}$, $\mathcal{L}_\text{pseudo}$, $\mathcal{L}_\text{TV}$ or $\mathcal{L}_\text{smooth}$, the output videos suffer from distorted geometric shapes, unsatisfactory flickering artifacts or temporal choppiness. In comparison, our full model delivers the best performance. 

\begin{table}[t]
\centering
\caption{Ablation study for our proposed components.}
\vspace{-2mm}
  \resizebox{\columnwidth}{!}{
\begin{tabular}{lccccccc}
\toprule 
Method & \multicolumn{1}{c}{CLIP$\downarrow$} & \multicolumn{1}{c}{CLIP-T-f$\downarrow$} & \multicolumn{1}{c}{CLIP-T-b$\downarrow$} & \multicolumn{1}{c}{CLIP-T-s$\downarrow$}   & \multicolumn{1}{c}{XCLIP-f$\uparrow$} & \multicolumn{1}{c}{XCLIP-b$\uparrow$} & \multicolumn{1}{c}{XCLIP-r$\uparrow$} \\
\midrule
w/o $\mathcal{L}_\text{recon}$    &0.2828  &0.0401  &0.0126  &{0.0401}& 14.8765 & 15.0583 & 13.7076\\
w/o $\mathcal{L}_\text{SDS}$   & 0.3044 & 0.0150 & 0.0188 & 0.0994 & 17.5322 & 13.1683 & 13.8437 \\
w/o $\mathcal{L}_\text{pseudo}$  & 0.3103 & 0.0137 & 0.0286 & 0.0978 & 16.8965 & 14.2132 & 10.8373\\

w/o $\mathcal{L}_\text{TV}$    &0.2603 &0.0091 & {0.0108}  &0.0158&17.2587 & 17.8444 & 16.6795\\
w/o $\mathcal{L}_\text{smooth}$  &0.2549  &0.0103 & {0.0148}  &0.0205& 16.3765 & 15.2553 & 16.3614\\

Full model                 & \textbf{0.2491} & \textbf{0.0072} &\textbf{0.0079} & \textbf{0.0068} &\textbf{17.5926}&\textbf{18.4030} & \textbf{16.8225}\\
\bottomrule
\end{tabular}
}
\vspace{-2mm}
\label{tab:abl}
\end{table}

\subsection{Limitation}
\looseness=-1
Despite our exciting 4D generation results, 4DGen still has some limitations. 
Multi-object generation is beyond our assumption since our diffusion prior is pre-trained on object-centric 3D datasets. We will extend our framework to compositional and scene-level generation in the future.

\section{Conclusion}
\label{sec:conclusion}

In this work, we construct a novel 4DGen pipeline for grounded 4D content creation. 
We empower users to fully control the appearance and motion by specifying a static 3D asset or a monocular video sequence.
Armed with our spatial-temporal pseudo labels and seamless consistency regularization, we effectively construct deformable 3D Gaussians that can be rendered from any viewpoint at any timestep. 
Our framework presents a faithful reconstruction of the input signals and visually pleasing novel view synthesis results at arbitrary timesteps.

\bibliographystyle{splncs04}
\bibliography{main}
\end{document}